\documentclass{templates/llncs} 

\usepackage{IEEEtrantools}

\usepackage{etoolbox}
\newcommand{\repthanks}[1]{\textsuperscript{\ref{#1}}}
\makeatletter
\patchcmd{\maketitle}
  {\def\thanks}
  {\let\repthanks\repthanksunskip\def\thanks}
  {}{}
\patchcmd{\@maketitle}
  {\def\thanks}
  {\let\repthanks\@gobble\def\thanks}
  {}{}
\newcommand\repthanksunskip[1]{\unskip{}}
\makeatother
\usepackage{hyperref}
\usepackage{physics}
\usepackage{graphicx}
\usepackage{gensymb}
\usepackage{booktabs}
\usepackage{siunitx}
\usepackage{floatrow}

\usepackage{IEEEtrantools}
\bstctlcite{IEEEexample:BSTcontrol}

\usepackage{multirow}
\newfloatcommand{capbtabbox}{table}[][\FBwidth]

\begin{document}

\bstctlcite{IEEEexample:BSTcontrol}

	\title{No New-Net}

  \author{Fabian Isensee \inst{1} \and
    Philipp Kickingereder \inst{2} \and
    Wolfgang Wick \inst{3} \and
    Martin Bendszus \inst{2} \and
    Klaus H. Maier-Hein \inst{1}}
  \authorrunning{Fabian Isensee et al.} 
  %
  \tocauthor{**************, **************, **************}
    \institute{Division of Medical Image Computing, German Cancer Research Center (DKFZ),\\ Heidelberg, Germany  
    \and Department of Neuroradiology, University of Heidelberg Medical Center,\\ Heidelberg, Germany
    \and Neurology Clinic, University of Heidelberg Medical Center,\\ Heidelberg, Germany
    }
	
	\maketitle   

	\begin{abstract}
	    In this paper we demonstrate the effectiveness of a well trained U-Net in the context of the BraTS 2018 challenge. This endeavour is particularly interesting given that researchers are currently besting each other with architectural modifications that are intended to improve the segmentation performance. We instead focus on the training process arguing that a well trained U-Net is hard to beat. Our baseline U-Net, which has only minor modifications and is trained with a large patch size and a Dice loss function indeed achieved competitive Dice scores on the BraTS2018 validation data. By incorporating additional measures such as region based training, additional training data, a simple postprocessing technique and a combination of loss functions, we obtain Dice scores of 77.88, 87.81 and 80.62, and Hausdorff Distances (95th percentile) of 2.90, 6.03 and 5.08 for the enhancing tumor, whole tumor and tumor core, respectively on the test data. This setup achieved rank two in BraTS2018, with more than 60 teams participating in the challenge.

\keywords{CNN, Brain Tumor, Glioblastoma, U-Net, Dice loss} 
\end{abstract}
	\section{Introduction}
	Quantitative assessment of brain tumors provides valuable information and therefore constitutes an essential part of diagnostic procedures. Automatic segmentation is attractive in this context, as it allows for faster, more objective and potentially more accurate description of relevant tumor parameters, such as the volume of its subregions. Due to the irregular nature of tumors, however, the development of algorithms capable of automatic segmentation remains challenging.
	
	The brain tumor segmentation challenge (BraTS) \cite{menze2015multimodal} aims at encouraging the development of state of the art methods for tumor segmentation by providing a large dataset of annotated low grade gliomas (LGG) and high grade glioblastomas (HGG). The BraTS 2018 training dataset, which consists of 210 HGG and 75 LGG cases, was annotated manually by one to four raters and all segmentations were approved by expert raters \cite{bakas_data,bakas_data_2,bakas_data_3}. For each patient a T1 weighted, a post-contrast T1-weighted, a T2-weighted and a Fluid-Attenuated Inversion Recovery (FLAIR) MRI was provided. The MRI originate from 19 institutions and were acquired with different protocols, magnetic field strengths and MRI scanners. Each tumor was segmented into edema, necrosis and non-enhancing tumor and active/enhancing tumor. The segmentation performance of participating algorithms is measured based on the DICE coefficient, sensitivity, specificity and 95th percentile of Hausdorff distance.
	
	It is unchallenged by now that convolutional neural networks (CNNs) dictate the state of the art in biomedical image segmentation \cite{kamnitsas2017efficient,isensee2017brain,li2017h,isensee2017automatic,kamnitsas2017ensembles,wang2017automatic}. As a consequence, all winning contributions to recent BraTS challenges were exclusively build around CNNs. One of the first notably successful neural network for brain tumor segmentation was DeepMedic, a 3D CNN introduced by Kamnitsas et al. \cite{kamnitsas2017efficient}. It comprises a low and a high resolution pathway that capture semantic information at different scales and recombines them to predict a segmentation based on precise local as well as global image information. Kamnitsas et al. later enhanced their architectures with residual connections for BraTS 2016 \cite{kamnitsas2016deepmedic}. 
	With the success of encoder-decoder architectures for semantic segmentation, such as FCN \cite{long2015fully,chen2018deeplab} and most notably the U-Net \cite{ronneberger2015u}, it is unsurprising that these architectures are used in the context of brain tumor segmentation as well. In BraTS 2017, all winning contributions were at least partially based on encoder-decoder networks. Kamnitsas et al. \cite{kamnitsas2017ensembles}, who were the clear winner of the challenge, created an ensemble by combining three different network architectures, namely 3D FCN \cite{long2015fully}, 3D U-Net \cite{cciccek20163d,ronneberger2015u} and DeepMedic \cite{kamnitsas2017efficient}, trained with different loss functions (Dice loss \cite{drozdzal2016importance,milletari2016v} and crossentropy) and different normalization schemes. Wang et al. \cite{wang2017automatic} used a FCN inspired architecture, enhanced with dilated convolutions \cite{chen2018deeplab} and residual connections \cite{he2016identity}. Instead of directly learning to predict the regions of interest, they trained a cascade of networks that would first segment the whole tumor, then given the whole tumor the tumor core and finally given the tumor core the enhancing tumor. Isensee et al. \cite{isensee2017brain} employed a U-Net inspired architecture that was trained on large input patches to allow the network to capture as much contextual information as possible. This architecture made use of residual connections \cite{he2016identity} in the encoder only, while keeping the decoder part of the network as simple as possible. The network was trained with a multiclass Dice loss and deep supervision to improve the gradient flow.
	
	Recently, a growing number of architectural modifications to encoder-decoder networks have been proposed that are designed to improve the performance of the networks for their specific tasks \cite{milletari2016v,jegou2017one,oktay2018attention,roy2018concurrent,wang2017automatic,isensee2017brain,kayalibay2017cnn,li2017h}. Due to the sheer number of such variants, it becomes increasingly difficult for researchers to keep track of which modifications extend their usefulness over the few datasets they are typically demonstrated on. We have implemented a number of these variants and found that they provide no additional benefit if integrated into a well trained U-Net. In this context, our contribution to the BraTS 2018 challenge is intended to demonstrate that such a U-Net, without using significant architectural alterations, is capable of generating competitive state of the art segmentations.

	\section{Methods}
	In the following we present the network architecture and training schemes used for our submission. As hinted in the previous paragraph, we will use a 3D U-Net architecture that is very close to its original publication \cite{cciccek20163d} and optimize the training procedure to maximize its performance on the BraTS 2018 training and validation data.
	
	\label{methods}	

	\subsection{Preprocessing}
	With MRI intensity values being non standardized, normalization is critical to allow for data from different institutes, scanners and acquired with varying protocols to be processed by one single algorithm. This is particularly true for neural networks where imaging modalities are typically treated as color channels. Here we need to ensure that the value ranges match not only between patients but between the modalities as well in order to avoid initial biases of the network. We found the following workflow to work well. We normalize each modality of each patient independently by subtracting the mean and dividing by the standard deviation of the brain region. The region outside the brain is set to 0. As opposed to normalizing the entire image including the background, this strategy will yield comparative intensity values within the brain region irrespective of the size of the background region around it.
	
	\subsection{Network architecture}

\begin{figure}[t!]
    	\begin{center}
    		\includegraphics[width=\textwidth]{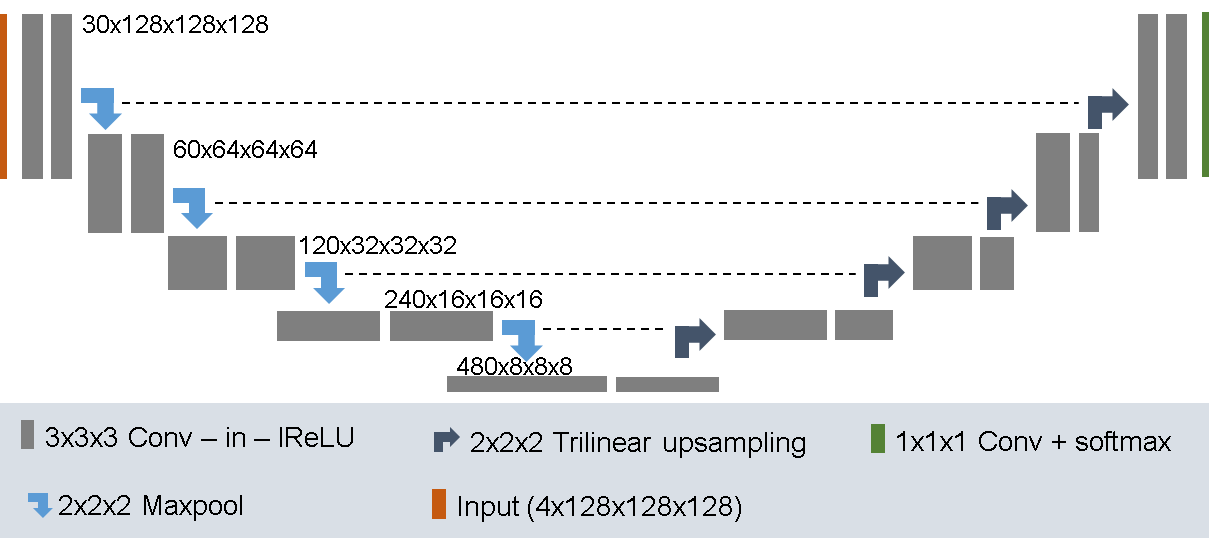}
    	\end{center}
      \caption{We use a 3D U-Net architecture with minor modifications. It uses instance normalization \cite{ulyanov2016instance} and leaky ReLU nonlinearities and reduces the number of feature maps before upsampling. Feature map dimensionality is noted next to the convolutional blocks, with the first number being the number of feature channels.}
	\label{fig:architecture}
    \end{figure}

	U-Net \cite{ronneberger2015u} is a successful encoder-decoder network that has received a lot of attention in the recent years. Its encoder part works similarly to a traditional classification CNN in that it successively aggregates semantic information at the expense of reduced spatial information. Since in segmentation, both semantic as well as spatial information are crucial for the success of a network, the missing spatial information must somehow be recovered. U-Net does this through the decoder, which receives semantic information from the bottom of the 'U' (see Fig. \ref{fig:architecture}) and recombines it with higher resolution feature maps obtained directly from the encoder through skip connections. Unlike other segmentation networks, such as FCN \cite{long2015fully} and previous iterations of DeepLab \cite{chen2018deeplab} this allows U-Net to segment fine structures particularly well.
	
	Our network architecture is an instantiation of the 3D U-Net \cite{cciccek20163d} with minor modifications. Following our successful participation in 2017 \cite{isensee2017brain}, we stick with our design choice to process patches of size 128x128x128 with a batch size of two. Due to the high memory consumption of 3D convolutions with large patch sizes, we implemented our network carefully to still allow for an adequate number of feature maps. By reducing the number of filters right before upsampling and by using inplace operations whenever possible, this results in a network with 30 feature channels at the highest resolution, which is nearly double the number we could train with in our previous model (using the same 12 GB NVIDIA Titan X GPU). Due to our choice of loss function, traditional ReLU activation functions did not reliably produce the desired results, which is why we replaced them with leaky ReLUs (leakiness $10^{-2}$) throughout the entire network. With a small batch size of 2, the exponential moving averages of mean and variance within a batch learned by batch normalization \cite{ioffe2015batch} are unstable and do not reflect the feature map activations at test time very well. We found instance normalization \cite{ulyanov2016instance} to provide more consistent results and therefore used it to normalize all feature map activations (between convolution and nonlinearity). For an overview over our segmentation architecture, please refer to Fig. \ref{fig:architecture}.

	\subsection{Training Procedure}
	\label{training_procedure}
	Our network architecture is trained with randomly sampled patches of size 128x128x128 voxels and batch size 2. We refer to an epoch as an iteration over 250 batches and train for a maximum of 500 epochs. The training is terminated early if the exponential moving average of the validation loss ($\alpha = 0.95$) has not improved within the last 60 epochs. Training is done using the ADAM optimizer with an initial learning rate $\mathrm{lr_{init}} = 1 \cdot 10^{-4}$, which is reduced by factor 5 whenever the above mentioned moving average of the validation loss has not improved in the last 30 epochs. We regularize with a l2 weight decay of $10^{-5}$.
	
	One of the main challenges with brain tumor segmentation is the class imbalance in the dataset. While networks will train with crossentropy loss function, the resulting segmentations may not be ideal in the sense of the Dice score they obtain. Since the Dice scores is one of the most important metrics based upon which contributions are ranked, it is imperative to optimize this metric. We achieve that by using a soft Dice loss for the training of our network. While several formulations of the Dice loss exist in the literature \cite{sudre2017generalised,drozdzal2016importance,milletari2016v}, we prefer to use a multi-class adaptation of \cite{drozdzal2016importance} which has given us good results in segmentation challenges in the past \cite{isensee2017automatic,isensee2017brain}. This multiclass Dice loss function is differentiable and can be easily integrated into deep learning frameworks:

	\begin{equation}
	  \mathcal{L}_\mathrm{dc} = - \frac{2}{|K|} \sum_{k\in K}\frac{\sum_i u_i^k v_i^k}{\sum_i u_i^k + \sum_i v_i^k}
	\end{equation}
	
	where $u$ is the softmax output of the network and $v$ is a one hot encoding of the ground truth segmentation map. Both $u$ and $v$ have shape $i$ by $c$ with $i$ being the number of pixels in the training patch and $k\in K$ being the classes.
	
  When training large neural networks from limited training data, special care has to be taken to prevent overfitting. We address this problem by utilizing a large variety of data augmentation techniques. The following augmentation techniques were applied on the fly during training: random rotations, random scaling, random elastic deformations, gamma correction augmentation and mirroring. Data augmentation was done with our own in-house framework which is publically available at \href{https://github.com/MIC-DKFZ/batchgenerators}{https://github.com/MIC-DKFZ/batchgenerators}.
  
  The fully convolutional nature of our network allows to process arbitrarily sized inputs. At test time we therefore segment an entire patient at once, alleviating problems that may arise when computing the segmentation in tiles with a network that has padded convolutions. We furthermore use test time data augmentation by mirroring the images and averaging the softmax outputs.
  
  \subsection{Region based prediction}
  Wang et al. \cite{wang2017automatic} use a cascade of CNNs to segment first the whole tumor, then the tumor core and finally the enhancing tumor. We believe the cascade and their rather complicated network architecture to be of lesser importance, but the fact that they did not learn the labels (enhancing tumor, edema, necrosis) but instead optimized the regions that are finally evaluated in the challenge directly to be key to their good performance in last years challenge. For this reason we will also train a version of our model where we replace the final softmax with a sigmoid and optimize the three (overlapping) regions (whole tumor, tumor core and enhancong tumor) directly with the Dice loss.
  
  \subsection{Cotraining}
  285 training cases is a lot for medical image segmentation, but may still not be enough to prevent overfitting entirely. We therefore also experiment with cotraining on additional public and institutional data. For public data, we chose to use the BraTS data made available in the context of the Medical Segmentation Decathlon (\href{http://medicaldecathlon.com}{http://medicaldecathlon.com}). This dataset comprises 484 cases with ground truth segmentations collected from older BraTS challenges. 
  
  Cotraining is done for only two datasets at a time. Given that the label definitions between BraTS 2018 and the other datasets may differ, we use separate segmentation layers (1x1x1 convolution) at the end, which act as a supervised version of m heads \cite{lee2015m}. During training, each segmentation layer only receives gradients from examples of its corresponding dataset. The losses of both layers are averaged to obtain the total loss of a minibatch. The rest of the network weights are shared.
  
  \subsection{Postprocessing}
  One of the most challenging parts in the BraTS challenge data is distinguishing small blood vessels in the tumor core region (that must be labeled either as edema of as necrosis) from enhancing tumor. This is particularly detrimental for LGG patients that may have no enhancing tumor at all. The BraTS challenge awards a Dice score of 1 if a label is absent in both the ground truth and the prediction. Conversely, only a single false positive voxel in a patient where no enhancing tumor is present in the ground truth will result in a Dice score of 0. Therefore we replace all enhancing tumor voxels with necrosis if the total number of predicted enhancing tumor is less than some threshold. This threshold is chosen for each experiment independently by optimizing the mean Dice (using the above mentioned convention) on the BraTS 2018 training cases.
  
  \subsection{Dice and Cross-entropy}
  While being widely popular and providing state of the art results on many medical segmentation challenges, the Dice loss has some downsides, such as badly calibrated softmax probabilities (basically binary 0-1 predictions) and occasional convergence issues (if the true positive term is too small for rare classes) compared to the negative log-likelihood loss (also referred to as cross-entroy loss function). We therefore also experiment with using these losses in conjunction by using both a Dice as well as a negative log-likelihood term and simply adding them together to form the total loss (unweighted sum).
  
\section{Experiments and Results}
  We designed our training scheme by running a five fold cross-validation on the 285 training cases of BraTS 2018. If additional data is used, the additional training cases are split into five folds as well and used for co-training. Training set results are summarized in Table \ref{tab:results}, validation set results can be found in table \ref{tab:val}. Unless noted otherwise, validation set results were obtained by using the five networks from the training cross-validation as an ensemble. For consistency with other publications, all reported values were computed by the online evaluation platform (\href{}{https://ipp.cbica.upenn.edu/}).
  
  Due to the relatively small size of the validation set (66 cases vs 285 training cases) we base our main analysis on the cross-validation results. We are confident that conclusions drawn from the training set are more robust and will generalize well to the test set.
  
\begin{table}[]
\label{tab:results}
\caption{Results on BraTS 2018 training data (285 cases). All results were obtained by running a five fold cross-validation. Metrics were computed by the online evaluation platform.}
\begin{tabular}{lccclll}
    & \multicolumn{3}{c}{Dice}     & \multicolumn{3}{c}{HD95}  \\
\multicolumn{1}{c}{}    & enh.  & whole & core  & enh. & whole   & core \\
\hline
Isensee et al. (2017) \cite{isensee2017brain}     & 70.69 & 89.51 & 82.76 & 6.24 & 6.04 & 6.95 \\
baseline     & 73.43 & 89.76 & 82.17 & 4.88 & 5.86 & 7.11 \\
baseline + reg    & 73.81 & 90.02 & 82.87 & 5.01 & 6.26 & 6.48 \\
baseline + reg + cotr (dec)  & 75.94 & 91.33 & 85.28 & 4.29 & 4.82 & 5.05 \\
baseline + reg + cotr (dec) + post & \textbf{78.68} & 91.33 & 85.28 & 3.49 & \textbf{4.82} & \textbf{5.05} \\
baseline + reg + cotr (dec) + post + DC\&CE & 78.62 & \textbf{91.75} & \textbf{85.69} & \textbf{2.84} & 4.88 & 5.11 \\
baseline + reg + cotr (inst) + post + DC\&CE & 76.32 & 90.35 & 84.36 & 3.74 & 5.64 & 5.98 \\
baseline + reg + post + DC\&CE & 76.78 & 90.30 & 83.55 & 3.66 & 5.36 & 6.03
\end{tabular}
\end{table}

\begin{table}[]
\caption{Results on BraTS2018 validation data (66 cases). Results were obtained by using the five models from the training set cross-validation as an ensemble. Metrics were computed by the online evaluation platform.}
\begin{tabular}{lccclll}
    & \multicolumn{3}{c}{Dice}     & \multicolumn{3}{c}{HD95}  \\
\multicolumn{1}{c}{}    & enh.  & whole & core  & enh. & whole   & core \\
\hline
baseline & 79.59 & 90.80 & 84.32 & 3.12 & 4.79 & 8.16 \\
baseline + reg + cotr (dec) + post + DC\&CE (*) & 80.46 & 91.21 & 85.77 & 2.52 & 4.38 & 6.73 \\
baseline + reg + cotr (inst) + post + DC\&CE (**) & 80.95 & 91.15 & 86.6 & 2.44 & 5.02 & 6.73\\
baseline + reg + post + DC\&CE & 80.66 & 90.92 & 85.22 & 2.74 & 5.83 & 7.20 \\
ensemble of (*) and (**) & 80.87 & 91.26 & 86.34 & 2.41 & 4.27 & 6.52
\end{tabular}
\label{tab:val}
\end{table}

  Results on the BraTS2018 training data are summarized in table \ref{tab:results}. We refer to our basic U-Net that was trained on BraTS2018 training data with large input patches and a Dice loss function as \textit{baseline}. With Dice scores of 73.43/89.76/82.17 (enh/whole/core) on the training set this baseline model is by itself already very strong, especially when compared to the model of Isensee et al. \cite{isensee2017brain} that achieved the third place in BraTS2017 (the training data for both challenges is identical, allowing a direct comparison of the models). Adding region based training (\textit{reg}) improved the Dice scores of both the enhancing tumor as well as the tumor core.  When training with decathlon data (\textit{cotr (dec)}), we gain two Dice points in enhancing tumor and minor improvements for the tumor core. Our postprocessing, which is targeted at correcting false positive enhancing tumor predictions in LGG patients has a substantial impact on enhancing tumor Dice. On the training set it increases the mean enhancing tumor Dice by almost three points. Using the sum Dice and cross-entropy as a loss function yields yet another small improvement. Interestingly, using our institutional data for cotraining yields much worse results on the training set. In order to isolate the impact of additional training data we added the model \textit{baseline + reg + post + DC$\&$CE} to the table.
  
  While the model that uses institutional data performed worse on the training set, it was slightly better on the validation set (see table \ref{tab:val}). We explain this discrepancy by the possibility that the Dice and Hausdorff distance scores obtained from the training set cross-validation may be overestimated when cotraining with decathlon data. Since any potential case correspondences between decathlon data and BraTS2018 is unknown due to the naming scheme of the decathlon cases, we cannot exclude the possibility that cases that are currently in the validation split for BraTS 2018 appear in the training split of the decathlon data (albeit with different ground truth segmentations). This uncertainty, along with the strong performance of the model cotrained with institutional data on the validation set led us to the decision to submit an ensemble of these two models. The ensemble achieves Dice scores of 80.87/91.26/86.34 (enh/whole/core) and Hausdorff distances of 2.41/4.27/6.52 on the validation set. For comparison, we also included the validation set result achieved with no additional training data. 
  
      \begin{figure}[t!]
	\begin{center}
		\includegraphics[width=\textwidth]{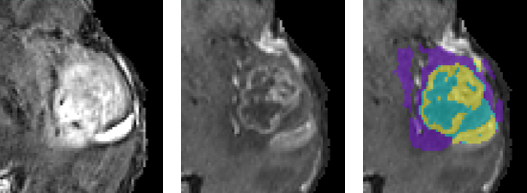}
	\end{center}
  \caption{Qualitative results. The case shown here is patient CBICA\_AZA\_1 from the validation set. Left: flair, middle: t1ce, right: our segmentation. Enhancing tumor is shown in yellow, necrosis in turquoise and edema in violet. }
	\label{fig:results}
\end{figure}

  Figure \ref{fig:results} shows a qualitative example segmentation. The patient shown is taken from the validation set (CBICA\_AZA\_1). As can be seen in the middle (t1ce), there are several blood vessels close to the enhancing tumor. Segmentation CNNs typically struggle to correctly differentiate between such vessels and actual enhancing tumor. This is most likely due to a) a difficulty in detecting tube-like structures b) few training cases where these vessels are an issue c) the use of Dice loss functions that does not sufficiently penalize false segmentations of vessels due to their relatively small size. In the case shown here, our model correctly segmented the vessels as background.

\begin{table}[]
\begin{tabular}{cc|cccccc}
 &  & \multicolumn{3}{c}{Dice} & \multicolumn{3}{c}{Hausd. dist.} \\
 &  & enh. & whole & core & enh. & whole & core \\ \hline
\multirow{3}{*}{NVDLMED} & Mean & 76.64 & 88.39 & 81.54 & 3.77 & 5.90 & 4.81 \\
 & StdDev & 25.57 & 11.83 & 24.99 & 8.61 & 10.01 & 7.52 \\
 & Median & 84.41 & 92.06 & 91.67 & 1.73 & 3.16 & 2.45 \\ \hline
\multirow{3}{*}{MIC-DKFZ} & Mean & 77.88 & 87.81 & 80.62 & 2.90 & 6.03 & 5.08 \\
 & StdDev & 23.93 & 12.89 & 25.02 & 3.85 & 9.98 & 8.09 \\
 & Median & 84.94 & 91.79 & 90.72 & 1.73 & 3.16 & 2.83 
\end{tabular}
\label{tab:test}
\caption{Test set results of NVDLMED, the winner of BraTS2018, and our method, which achieved the second place.}
\end{table}

Test set results (as communicated by the organizers of the challenge) are presented in table \ref{tab:test}. We used used an ensemble of the two models that were trained with institutional and decathlon data for our final submission. Each of these models is in turn an ensemble of five models resulting from the corresponding cross-validation, resulting in a total of 10 predictions for each test case. Our algorithm achieved the second place out of 64 participating teams. We compare our results to the winning contribution by Myronenko et al. (team NVDLMED). While our model had strong results for enhancing tumor, NVDLMED outperformed our approach in both tumor core and whole tumor. Please refer to \cite{bakas2018identifying} for a detailed summary of the challenge results.

\section{Discussion}
  In this paper we demonstrated that a generic U-Net architecture that has only minor modifications can obtain very competitive segmentation, if trained correctly. While our base model is already quite strong, enhancing its training procedure by using region-based training, cotraining with additional training data, postprocessing to target false positive enhancing tumor detection as well as a combination of Dice and cross-entropy loss, increases its performance substantially. For our final submission we used an ensemble of a model cotrained with public and another cotrained with institutional data. Despite using only a generic U-Net architecture, our approach achieved the second place in the BraTS2018 challenge, underligning the impact a well designed framework can have on model training.

\bibliographystyle{IEEEtran}
\bibliography{bibliography}
\end{document}